\documentclass[letterpaper]{article} 
\usepackage{aaai23}  
\usepackage{times}  
\usepackage{helvet}  
\usepackage{courier}  
\usepackage[hyphens]{url}  
\usepackage{graphicx} 
\usepackage{natbib}  
\usepackage{caption} 
\usepackage{algorithm}
\usepackage{algorithmic}
\usepackage{color,xcolor}
\usepackage{array}
\usepackage{amsmath,multicol,multirow}
\usepackage{amsthm,amsfonts}
\usepackage{booktabs}
\usepackage{caption}
\usepackage{graphicx}
\usepackage{float}
\usepackage{subcaption}
\usepackage{newfloat}
\usepackage{listings}
\DeclareCaptionStyle{ruled}{labelfont=normalfont,labelsep=colon,strut=off} 
\floatstyle{ruled}
\newfloat{listing}{tb}{lst}{}
\floatname{listing}{Listing}
\title{Semi-supervised Credit Card Fraud Detection via \\ Attribute-Driven Graph Representation}
\author{
    Sheng Xiang\textsuperscript{\rm 1},
    Mingzhi Zhu\textsuperscript{\rm 2},
    Dawei Cheng\textsuperscript{\rm 2,3}\thanks{Corresponding Author is Dawei Cheng.},
    Enxia Li\textsuperscript{\rm 1},\\
    Ruihui Zhao\textsuperscript{\rm 4},
    Yi Ouyang\textsuperscript{\rm 4},
    Ling Chen\textsuperscript{\rm 1},
    Yefeng Zheng\textsuperscript{\rm 4}
}
\affiliations{
    \textsuperscript{\rm 1}Australian Artificial Intelligence Institute, University of Technology Sydney, Sydney, Australia\\
    \textsuperscript{\rm 2}Department of Computer Science and Technology, Tongji University, Shanghai, China\\
    \textsuperscript{\rm 3}Shanghai Artificial Intelligence Laboratory, Shanghai, China\\
    \textsuperscript{\rm 4}Tencent Jarvis Laboratory, Shenzhen, China\\

    \{sheng.xiang, ling.chen\}@uts.edu.au, mz3379@nyu.edu, dcheng@tongji.edu.cn, \\
    enxia.li@student.uts.edu.au, zachary@ruri.waseda.jp, \{yiouyang, yefengzheng\}@tencent.com
}

\usepackage{bibentry}

\begin{document}

\maketitle

\begin{abstract}
Credit card fraud incurs a considerable cost for both cardholders and issuing banks. Contemporary methods apply machine learning-based classifiers to detect fraudulent behavior from labeled transaction records. But labeled data are usually a small proportion of billions of real transactions due to expensive labeling costs, which implies that they do not well exploit many natural features from unlabeled data. Therefore, we propose a semi-supervised graph neural network for fraud detection. Specifically, we leverage transaction records to construct a temporal transaction graph, which is composed of temporal transactions (nodes) and interactions (edges) among them. Then we pass messages among the nodes through a Gated Temporal Attention Network (GTAN) to learn the transaction representation. We further model the fraud patterns through risk propagation among transactions. The extensive experiments are conducted on a real-world transaction dataset and two publicly available fraud detection datasets. The result shows that our proposed method, namely GTAN, outperforms other state-of-the-art baselines on three fraud detection datasets. Semi-supervised experiments demonstrate the excellent fraud detection performance of our model with only a tiny proportion of labeled data.
\end{abstract}

\section{Introduction}
The great losses caused by financial fraud have attracted continuous attention from academia, industry, and regulatory agencies.
For instance, as reported in~\cite{alfalahi2019conceptual,mate2019effects}, financial fraud detection plays a critical role to support the sustainable economic growth.
However, fraudulent behaviors against online payments, such as illegal card swiping, have caused property losses to online payment users~\cite{Bhattacharyya2011DataMF}.

\begin{figure}[tb!]\vspace{-0pt}
  \centering
  \includegraphics[width=0.99\linewidth]{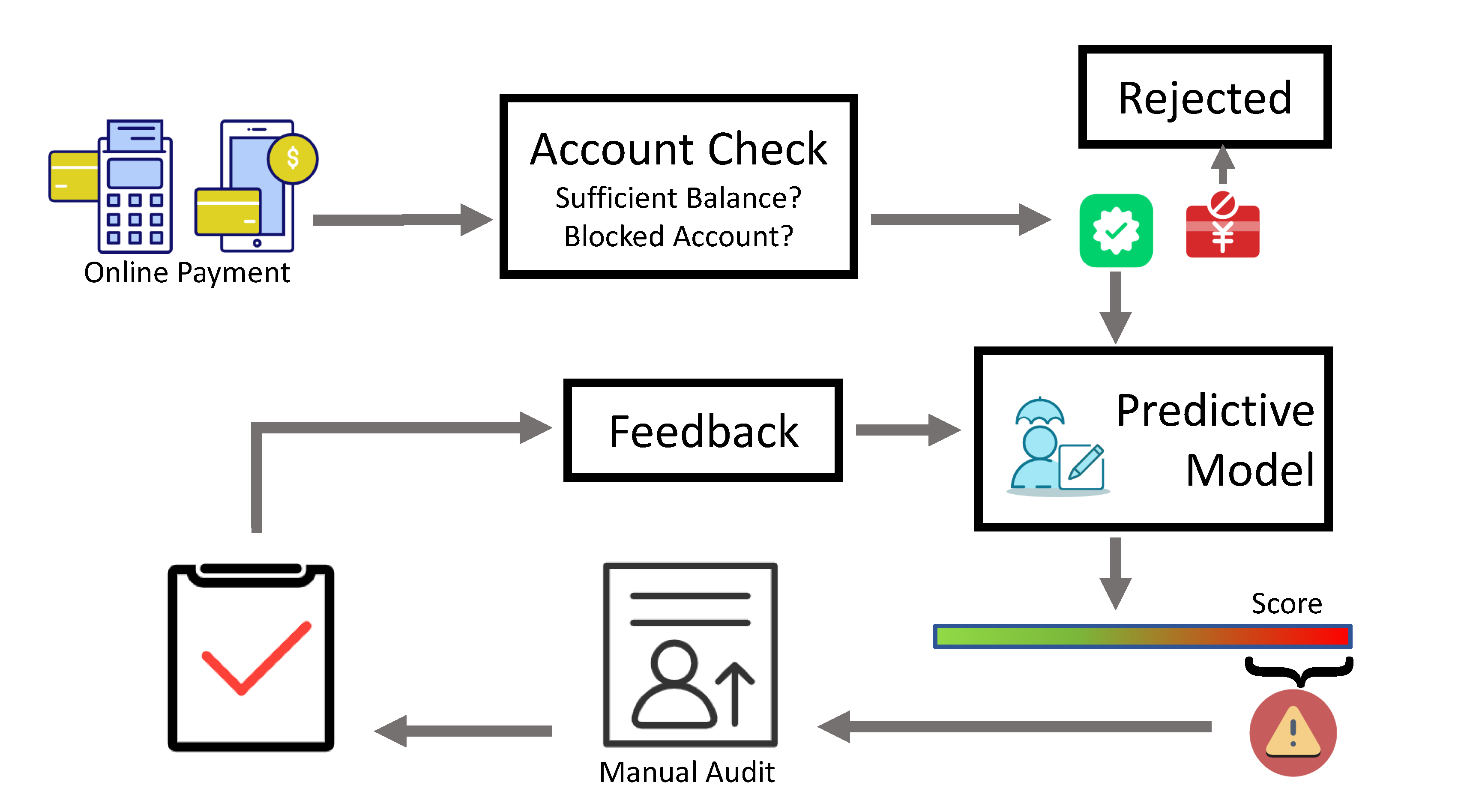}\vspace{-0pt}
  \caption{The illustration of typical credit card fraud detection process. The detection system of card issuer assesses each transaction with an online predictive model once it has passed account checking. } \label{fig:intro}\vspace{-5pt}
\end{figure}

An important line of research in financial fraud detection is credit card fraud detection, where credit card fraud is a general term for the unauthorized use of funds in a transaction, typically by means of a credit or debit card~\cite{bhattacharyya2011data}.
Figure \ref{fig:intro} shows a typical fraud detection framework deployed in the commercial system~\cite{cheng2020graph}.
A direct way to detect fraud is to match each transaction according to specific rules
such as card blacklists and budget checking. However, criminals will also obtain the knowledge of vulnerabilities from the response of the pre-designed rule system, thus invalidating the original system. To solve the invalidation problem, the predictive model is designed to automatically detect fraud patterns and produces a fraud risk score. Domain experts then can thereby focus on the high-risk transactions.

\vspace{1mm}
\noindent \textbf{State-of-the-art.}
In the literature, many existing predictive models have been extensively studied to deal with fraud transactions  (e.g.,~\cite{patidar2011credit,fu2016credit}),
which can be classified into two categories: (1) \textit{Rule-based methods} directly generate sophisticated rules by domain experts
to identify the suspicious transactions. For instance, authors in~\cite{seeja2014fraudminer} proposed an association rule method for mining frequent fraud rules; (2) \textit{Machine learning-based methods} learn static models by exploring large amounts of historical data.
For example, authors in~\cite{fiore2017using} extracted features based on neural networks and built supervised classifiers for detecting fraudulent transactions.
Recently, graph machine learning-based methods have been proposed~\cite{wang2019semi}
where the transactions are modeled as a graph, and the advanced graph embedding techniques are deployed.

\vspace{1mm}
\noindent \textbf{Motivation.}
The state-of-the-art fraud detection techniques \cite{dou2020enhancing,liu2020alleviating,liu2021pick} can well capture the temporal or graph-based patterns of the transactions and significantly advance the performance of credit card fraud detection. However, they have at least one of the following three main limitations:
(1) ignoring unlabeled data containing rich fraud pattern information; (2) ignoring categorical attributes, which are ubiquitous in the real production environment; and (3) requiring too much time on feature engineering, especially for categorical features.

These motivate us to design a semi-supervised graph neural network for credit card fraud detection. In particular, to capture the relationships among the credit card transactions associated with temporal information, we leverage a temporal transaction graph to model the time-relevant patterns. Besides, labeling the transactions is time-consuming and cost expensive. Only a tiny proportion (much less than 10\%) of transactions are labeled in billions of real-life transactions, which contains many fraud patterns that have not been detected. Therefore, it is crucial to exploit the natural features from unlabeled data. In this paper, we design a Gated Temporal Attention Network (GTAN) for the temporal transaction graph, which can extract temporal fraud patterns and exploit both labeled and unlabeled data. In addition, categorical attributes are ubiquitous and useful in real applications. Therefore, it is necessary to leverage useful information through an attribute-driven model. In this paper, we introduce an attribute learning layer for preprocessing the transaction attributes and add risk embedding as new categorical attributes, which can better model fraud patterns (e.g., attribute embedding learning and risk propagation).

\noindent \textbf{Contributions} of our work are summarized as follows:
\begin{itemize}
  \item We model credit card behaviors as a temporal transaction graph and formulate a credit card fraud detection problem as a semi-supervised node classification task.
  \item We present a novel attribute-driven temporal graph neural network for credit card fraud detection. Specifically, we propose a gated temporal attention network to extract temporal and attribute information. And we pass attributes and risk information on the temporal transaction graph to exploit both labeled and unlabeled data.
  \item Extensive experiments on three datasets show the superiority of our proposed GTAN on fraud detection. Semi-supervised experiment results show that, when leveraging rich information from the unlabeled data and a bit of labeled data, our proposed method detects more fraud transactions than baselines.
\end{itemize}

\begin{figure*}
\centerline{\includegraphics[width=1.0\linewidth,height=170pt]{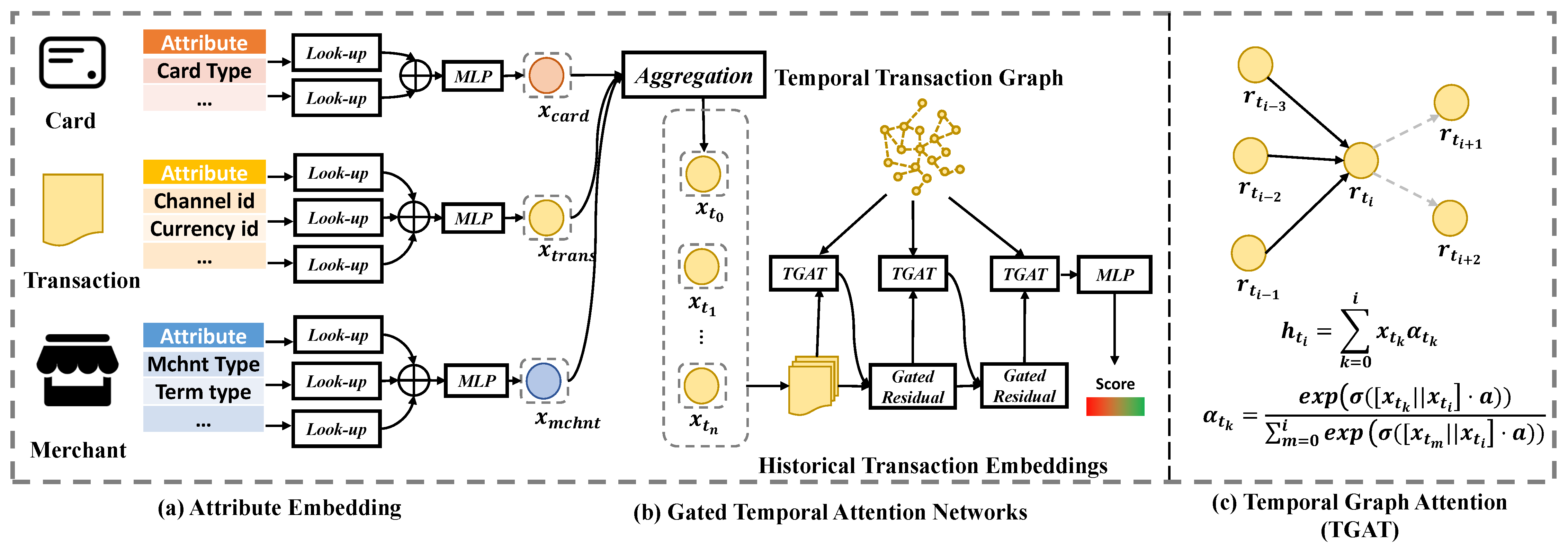}}
\caption{The illustration of the proposed model architecture and temporal graph attention mechanism.}\label{fig:modelarc}
\end{figure*}

\section{Related Works}
\label{sec:rel}

\subsection{Credit Card Fraud Detection}
Several machine learning techniques have been proposed in the literature to address the credit card fraud detection problem.
For instance, in~\cite{maes2002credit}, Bayesian Belief Networks (BBN) and Artificial Neural Networks (ANN) were applied on a real dataset obtained from Europay International.
In~\cite{csahin2011detecting} decision trees and support vector machines (SVMs) are applied on a real-world national bank dataset.
The authors in~\cite{fu2016credit} showed that using convolution to extract patterns can achieve higher accuracy than non-convolution neural networks. Recently, graph-based fraud detection techniques were proposed. For instance, CARE-GNN~\cite{dou2020enhancing} was proposed to tackle fraud detection on relational graphs. PC-GNN~\cite{liu2021pick} was proposed for imbalanced supervised learning on graphs.
\cite{fiore2017using} also proposed a generative adversarial network to improve the classification performance.
In~\cite{cheng2020graph,DBLP:conf/aaai/ChengXSZY020}, authors proposed joint feature learning based on spatial and temporal patterns.
However, they modeled the fraud patterns by using only one transaction/cardholder, thereby not being able to exploit the unlabeled data in real-life credit card transactions.
The approach we present in this paper is radically different, as we employ a semi-supervised architecture, where the fraud patterns on both unlabeled and labeled data are jointly learned within an attribute-driven graph neural network framework.

\subsection{Graph-Based Semi-supervised Learning}
Many recent works have shown the benefit of using unlabeled node attributes in graph neural networks for a wide range of prediction tasks~\cite{vaswani2017attention,Song2022GraphbasedSL}, such as text classification~\cite{xu2018structured}, molecule property prediction~\cite{guan2018diagnose} or language understanding~\cite{shen2018disan}. For instance, graph convolutional networks (GCN) were employed on partially labeled citation networks for node property prediction~\cite{kipf2016semi}.
GraphSAGE~\cite{hamilton2017inductive} was proposed to generate low-dimensional embeddings for previously unseen data. Graph attentive network model and random walks~\cite{wang2019semi} were deployed on social graphs to link the unlabeled and labeled data and pass messages among them.
However, they still face at least one of the following two limitations: (1) cannot scale up to real-world graphs over millions of nodes (e.g., vanilla graph attention networks~\cite{Velickovic2018GraphAN} have a space complexity $O(N^2)$, where $N$ denotes the number of nodes, which is unaffordable for tasks with millions of nodes); (2) cannot propagate and learn the categorical attribute embeddings, especially for risk embeddings.
Differently, our approach addresses the fraud detection problem via a message-passing model using categorical attributes, including risk information from node neighbors. Our work exploits attribute-driven model and semi-supervised graph neural networks to find more fraud patterns, which significantly improve the accuracy of credit card fraud detection.

\section{Proposed Method}
\label{sec:method}

In this section, we first introduce the framework of our proposed GTAN. Then, we present the process of feature engineering, the gated temporal attention networks, risk embedding and propagation, and the fraud detection classifier. Lastly, we introduce the optimization strategy.

\subsection{Model Architecture}

The general model architecture of our proposed method is illustrated in Figure~\ref{fig:modelarc}. Raw attributes of transaction records are first learned by the attribute embedding look-up and feature learning layer, which includes feature aggregation with a multi-layer perception (MLP). In our implementation, the attributes of the card include the card type, cardholder type, card limit, remaining limit, etc. The transaction attributes include the channel ID, currency ID, transaction amount, etc. The merchant attributes contain merchant type, terminal type, merchant location, sector, charge ratio, etc.
Then, we devise a gated temporal attention network to aggregate and learn the importance of historical transaction embeddings. Afterward, we leverage a two-layer MLP to learn the fraud probability from these representations. The whole model can be optimized in an end-to-end mechanism jointly with the existing stochastic gradient descent algorithm.

\subsection{Attribute Embedding and Feature Learning}
\label{subsec:attr}
Given transaction records $\mathbf{r}=(r_{1},r_{2},\cdots,r_{N})$, each record $r_i$ contains card attributes $f_c^i$, transaction attributes $f_r^i$, and merchant attributes $f_m^i$ as $r_i = {f_c^i, f_r^i, f_m^i}$. In preprocessing, different from~\cite{DBLP:conf/aaai/ChengXSZY020}, we do not filter out any cards or merchants with few authorized transaction records. As the number of cards and merchants which have never been checked manually is much larger than checked, we adopt full transaction records of users to maintain all potential frauds.
Afterward, we construct the numerical attribute representation of each record into tensor format $\mathbf{X}_{num} \in \mathbb{R}^{N\times d}$, where $N$ denotes the number of transactions, and $d$ denotes the dimensions of features. Besides, we extract the card, transaction, and merchant category attributes $\mathbf{X}_{cat} \in \mathbb{R}^{N\times d}$ separately through attribute embedding layers, which can be formulated as follows:
\begin{equation}
    \begin{split}
        e_{attr}=&\text{onehot}(f_{attr})\odot\mathbf{E}_{attr},\\
        x_{cat,i}=&\text{MLP}_i(\sum_{\forall j \in table_{i}}e_{j}), i\in\{card, trans, mchnt\},\\
    \end{split}
\end{equation}
where $j\in table_i$ denotes the column $j$ in our input table data $i$, $e_{attr}\in \mathbb{R}^{1\times d}$ denotes the embedding of attribute $attr$, $\text{onehot}(\cdot)$ denotes the one-hot encoding, $f_{attr}$ denotes the single attribute of one transaction, and $\mathbf{E}_{attr}\in \mathbb{R}^{m\times d}$ denotes the embedding matrix of attribute $attr$, where $m$ denotes the maximum number of attribute $attr$.
After obtaining the embedding vector of each attribute in the card, transaction, and merchant tables, we aggregate these embeddings to obtain each transaction's categorical embedding through add-pooling with $x_{cat}^{(u)}=\sum_ix_{cat,i}^{(u)}, i\in\{card, trans, mchnt\}$, where $x_{cat}^{(u)}\in \mathbb{R}^{1\times d}$ denotes the category embedding vector of the $u$-th transaction record.
To address the heterogeneity of categorical attributes, our proposed feature learning layer can model all categorical attributes and project them to a unified spatial dimension, which benefits our attribute-driven graph learning model.

\subsection{Gated Temporal Attention Networks}
To learn the temporal fraud patterns, we generate the temporal transaction graph~\cite{Xiang2021GeneralGG,Xiang2022EfficientLC} and aggregate messages on this graph to update the embedding of each transaction. Particularly, the directed temporal edges are generated with the previous transactions as the source and the current ones as the target, as illustrated in Figure~\ref{fig:modelarc}(c). Then we aggregate messages through Temporal Graph Attention (TGAT) mechanism~\cite{Xiang2022TemporalAH}. The number of generated temporal edges per node is a hyper-parameter, which will be studied in the experiment section.

\vspace{1mm}
\noindent \textbf{Temporal Graph Attention.} After the feature engineering and attribute embedding, we leverage a series of transaction embeddings $\mathbf{X}=\{x_{t_0},x_{t_1},...x_{t_n}\}$ to learn the temporal embedding of each transaction record. First, we combine categorical attributes and numerical attributes as the input of GTAN network with $x_{t_i} = x_{num}^{(t_i)}+x_{cat}^{(t_i)}$. At the first TGAT layer, we set $\mathbf{H}_0=\mathbf{X}$ as the input embedding matrix. Afterward, we leverage multi-head attention to separately calculate the importance of each neighbor and update embeddings, which can be formulated as follows:
\begin{equation}
    \begin{split}
        \mathbf{H}=&\text{Concat}(\text{Head}_1,...,\text{Head}_{h_{att}})\textbf{W}_{o},
    \end{split}
\end{equation}
where $h_{att}$ denotes the number of heads, $\textbf{W}_{o}\in \mathbb{R}^{d\times d}$ denotes learnable parameters, $\mathbf{H}$ denotes the aggregated embeddings with $\mathbf{H}=\{h_{t_0},h_{t_1},...,h_{t_n}\}$, and each attention head is formulated as follows:
\begin{equation}
    \begin{split}
        \text{Head}=&\sum_{x_i\in\mathcal{X}}\sigma(\sum_{x_t\in \mathcal{N}(x_i)}\alpha_{x_t,x_i}x_{t}),\\
        \alpha_{x_t,x_i}=&\frac{\text{exp}(\text{LeakyReLU}(\mathbf{a}^T[x_t||x_i))}{\sum_{x_j\in \mathcal{N}(x_t)}\text{exp}(\text{LeakyReLU}(\mathbf{a}^T[x_t||x_j]))},\\
    \end{split}
\end{equation}
where $\mathcal{N}(x_i)$ denotes the temporal neighbors of the $i$-th transaction, $\alpha_{x_t,x_i}$ denotes the importance of temporal edge $(x_t,x_i)$ in each attention head, and $\mathbf{a}\in\mathbb{R}^{2d}$ denotes the weight vector of each head. In practice, to avoid extra space consumption in extreme cases (such as high-frequency transactions in a short period), we use a neighbor sampling and truncation strategy to control the number of neighbor nodes $|\mathcal{N}(x_t)|$ (i.e., the number of associated temporal edges per node) through which the temporal graph attention layer propagates messages. Besides, to avoid borrowing future information, the neighbor transactions sampled for each transaction must be the past transactions from the same cardholder so that we can model the temporal fraud pattern through message passing on the temporal transaction graph.

\vspace{1mm}
\noindent \textbf{Attribute-driven Gated Residual.} To further improve the effectiveness and interpretability of our method, after obtaining aggregated embeddings, we leverage the embeddings and raw attributes to infer the importances of the aggregated embeddings and raw attributes after each layer of TGAT, which can be formulated as follows:
\begin{equation}
    \begin{split}
        \text{gate}_{t_i}=&\sigma([x_{cat,t_i}||x_{num,t_i}||h_{t_i}]\beta_{t_i}),\\
        z_{t_i}=&\text{gate}_{t_i}\cdot h_{t_i} + (1-\text{gate}_{t_i})\cdot x_{t_i},
    \end{split}
\end{equation}
where $\text{gate}_{t_i}\in [0,1]$ denotes the gate variable of the $t_i$-th transaction, $\sigma$ denotes the sigmoid activation function, $\beta_{t_i}\in \mathbb{R}^{3d\times 1}$ denotes the gate vector, and $z_{t_i}$ denotes the output vector of each TGAT layer, which is fed into the next layer as input. According to our framework, if we stack a new TGAT layer with the attribute-driven gated residual mechanism, we use the output of the $k$-th gating mechanism as the input of the $k+1$-th TGAT.

\subsection{Risk Embedding and Propagation}
Inspired by unifying label propagation with feature propagation~\cite{Shi2021MaskedLP}, we propose to take the manually annotated label as one of the categorical attributes of the transaction, and get the embedding of this categorical attribute, which we call \textit{risk embedding}. Specifically, we take the manually annotated label as the risk feature of each transaction, where the category of unlabeled data is `unlabeled', and the category of the rest of the data is `fraud' or `legitimate'. Then, we add this feature to the transaction data as one of our input categorical attributes. Due to concerns about label leakage, this attribute has not been used in previous fraud detection solutions. We will discuss the techniques for avoiding label leakage later. Then, we propose to embed the partially observed risk attributes (i.e., labels) into the same space as the other node features, which consist of the risk embedding vectors for labeled nodes and zero embedding vectors for the unlabeled ones. Then, we add the node features and risk embeddings together as input node features with $x_{t_i} = x_{num}^{(t_i)}+x_{cat}^{(t_i)}+\tilde{y}^{(t_1)}\mathbf{W}_r$, where $\mathbf{W}_r$ denotes the learnable parameters of risk embedding.
\cite{Shi2021MaskedLP} have proved that by mapping partially-labeled $\mathbf{\hat{Y}}$ and node features $\mathbf{X}$ into the same space and adding them up, we can use one graph neural network to achieve both attribute propagation and label propagation. Therefore, our fraud detection model can jointly model the temporal fraud patterns and risk propagation by adding the transaction label as one of the transaction categorical attributes.

\subsection{Fraud Risk Prediction}
After obtaining the aggregated embeddings of transactions, we leverage a two-layer MLP to predict the fraud risk, which is formulated as follows:
\begin{equation}
    \begin{split}
        \mathbf{\hat{y}}=\sigma(\text{PReLU}(\mathbf{H}\mathbf{W}_0+\mathbf{b}_0)\mathbf{W}_1+\mathbf{b}_1),
    \end{split}
\end{equation}
where $\mathbf{\hat{y}}\in \mathbb{R}^{N\times 1}$ denotes the risk prediction results of all transactions, and $\mathbf{W}$ and $\mathbf{b}$ denote the learnable parameters of MLP. Afterward, we calculate the objective function $\mathcal{L}$ via binary cross-entropy, which is formulated as follows:
\begin{equation}
    \begin{split}
        \mathcal{L}=-\frac{1}{N}\sum_{i=0}^{N}[&\mathbf{y}_i\cdot\log p(\mathbf{\hat{y}}_i|\mathbf{X},\mathbf{A})+\\
        &(1-\mathbf{y}_i)\cdot\log (1-p(\mathbf{\hat{y}}_i|\mathbf{X},\mathbf{A}))],
    \end{split}
\end{equation}
where $\mathbf{y}$ denotes the ground-truth label of transactions.
The proposed GTAN can be optimized through the standard SGD-based algorithms.

\subsection{Masking to Avoid Label Leakage}
\label{subsec:mask}
Previous works only took risk information as optimization objectives to supervise their fraud detection model training.
Unlike previous credit card fraud detection solutions, we semi-supervise our model by propagating transaction attributes and risk embeddings among labeled and unlabeled transactions. Using an unmasked objective for our fraud detection model will result in label leakage in the training process. In this case, our model will directly take observed labels and neglect the complicated hidden fraud patterns, which cannot be generalized in predicting future fraud transactions. Therefore, we propose to learn from the risk information of each transaction's neighbor transactions instead of learning from the label of itself. Specifically, a masked fraud detection training strategy is leveraged. During each training step, we randomly sample a batch of nodes, namely center nodes, along with the neighbor nodes corresponding to each center node. Then, we convert the partially observed labels $\mathbf{\hat{Y}}$ into $\mathbf{\Tilde{Y}}$ by masking all the center nodes' risk embeddings to zero embeddings and keeping the others unchanged. Then, our objective function is to predict $\mathbf{\hat{Y}}$ with given $\mathbf{X}$, $\mathbf{\Tilde{Y}}$ and $\mathbf{A}$:
\begin{equation}
    \begin{split}
        \mathcal{L}=-\frac{1}{|V|}\sum_{i=0}^{|V|}[&\mathbf{y}_i\cdot\log p(\mathbf{\hat{y}}_i|\mathbf{X},\mathbf{\tilde{Y}},\mathbf{A})+\\
        (1-&\mathbf{y}_i)\cdot\log(1-p(\mathbf{\hat{y}}_i|\mathbf{X},\mathbf{\tilde{Y}},\mathbf{A}))],
    \end{split}
\end{equation}
where $|V|$ represents the number of center nodes with masked labels. In this way, we can train our model without the self-loop leakage of risk information; and during inference, we employ all observed labels $\mathbf{\hat{Y}}$ as input categorical attributes to predict the risk of the transactions out of the training set. The optimization objective of our model can be intuitively summarized as modeling the fraud patterns by the attribute information of neighboring transaction nodes and the attribute information.

\section{Experiments}
\label{sec:exp}

In this section, we first describe the datasets used in the experiments, then compare our fraud detection performance with other state-of-the-art baselines on two supervised graph-based fraud detection datasets and one semi-supervised dataset. Then, we perform ablation studies by evaluating two variants of the proposed GTAN, which demonstrates the effectiveness of our proposed method and attribute-driven mechanism.

\subsection{Experiment Settings}
\subsubsection{Datasets}
To the best of our knowledge, we did not find any public semi-supervised credit card fraud detection dataset. Therefore, we collect the partially labeled records from our collaborated partners, namely Finacial Fraud Semi-supervised Dataset (\textbf{FFSD}). The ground truth labels are obtained on cases reported by consumers and confirmed by financial domain experts. If a transaction is reported by a cardholder or identified by financial experts as fraudulent, we label it as $1$; otherwise, it is labeled as $0$. Besides, we also experimented on two public supervised fraud detection datasets. The \textbf{YelpChi} graph dataset \cite{rayana2015collective} contains a selection of hotel and restaurant reviews on Yelp. Nodes in the graph of the YelpChi dataset are reviews with 32-dimensional features, and the edges are the relationships among reviews. The \textbf{Amazon} graph dataset \cite{mcauley2013amateurs} includes product reviews of musical instruments. The nodes in the graph are users with 25-dimensional features, and the edges are the relationships among reviews.
Some basic statistics of the three datasets are shown in Table~\ref{tab:data1}.

\begin{table}[!t]
\centering\vspace{-3pt}
 \tabcolsep 8pt
  \begin{tabular}{cccc}
    \toprule
    Dataset & YelpChi & Amazon & FFSD \\  \midrule
    \#Node & 45,954 & 11,948 & 1,820,840 \\
    \#Edge & 7,739,912 & 8,808,728 & 31,619,440 \\
    \#Fraud & 6,677 & 821 & 33,858 \\
    \#Legitimate & 39,277 & 11,127 & 141,861 \\
    \#Unlabeled & - & - & 1,645,121 \\
  \bottomrule
\end{tabular}
\caption{Statistics of the three fraud detection datasets.}  \label{tab:data1}
\vspace{-0pt}
\end{table}

\begin{table*}[t]\vspace{-0pt}\vspace{-3pt}
   \centering
   \tabcolsep 7pt
      \begin{tabular}{l@{\hspace{3\tabcolsep}}cccc@{\hspace{3\tabcolsep}}cccc@{\hspace{3\tabcolsep}}cccc}
         \toprule
         \multirow{2}{*}{Dataset} & \multicolumn{3}{c}{YelpChi} &  \multicolumn{3}{c}{Amazon} &  \multicolumn{3}{c}{FFSD}  \\
         \cmidrule(lr){2-4}\cmidrule(lr){5-7}\cmidrule(lr){8-10} & \multicolumn{1}{c}{AUC} & \multicolumn{1}{c}{F1} & \multicolumn{1}{c}{AP} & \multicolumn{1}{c}{AUC} & \multicolumn{1}{c}{F1} & \multicolumn{1}{c}{AP} & \multicolumn{1}{c}{AUC} & \multicolumn{1}{c}{F1} & \multicolumn{1}{c}{AP}    \\
         \midrule
         GEM & 0.5270 & 0.1060 & 0.1807 & 0.5261 & 0.0941 & 0.1159 & 0.5383 & 0.1490 & 0.1889   \\
         Player2Vec & 0.7003 & 0.4121&0.2473 &0.6185 & 0.2451&0.1291& 0.5278&0.2147&0.2041   \\
         FdGars & 0.7332&0.4420&0.2709 & 0.6556& 0.2713& 0.1438 & 0.6965&0.4089&0.2449   \\
         Semi-GNN & 0.5161 &0.1023 & 0.1811 & 0.7063 & 0.5492 &0.2254 & 0.5473&0.4485&0.2758\\
         GraphSAGE & 0.5364 & 0.4508 & 0.1712& 0.7502&  0.5795& 0.2624 & 0.6527 &0.5370 & 0.3844 \\
         GraphConsis & 0.7060 & 0.6041 & 0.3331 & 0.8782 & 0.7819 & 0.7336 & 0.6579 & 0.5466 & 0.3876   \\
         CARE-GNN & 0.7934&0.6493 &  0.4268 & 0.9115 & 0.8531 & 0.8219 & 0.6623 &0.5771 &0.4060  \\
         PC-GNN & 0.8174 & 0.6682 & 0.4810 & 0.9581 & 0.9153 & 0.8549 & 0.6795 & 0.6077 & 0.4487  \\

         \midrule
         GTAN &  \textbf{0.9241*}&	\textbf{0.7988*} &\textbf{0.7513*} & \textbf{0.9630*} & \textbf{0.9213*} & \textbf{0.8838*} & \textbf{0.7616*} &\textbf{0.6764*}&\textbf{0.5767*}   \\
         \bottomrule
   \end{tabular}
   \caption{Fraud detection performance on three datasets, compared with popular benchmark methods. We report the results of the area under the roc curve (AUC), macro average of F1 score (F1-macro), and averaged precision (AP).}\label{tab:result}
   \vspace{-5pt}
\end{table*}

\subsubsection{Compared Methods.}
The following methods are compared to highlight the effectiveness of the proposed GTAN.
\begin{itemize}
   \item \textit{GEM.} Heterogeneous GNN-based model proposed in~\cite{liu2018heterogeneous}. We set the learning rate to 0.1 and the number of hops of neighbors to 5.
   \item \textit{FdGars.} Fraudster detection via the graph convolutional networks proposed in~\cite{wang2019fdgars}. We set the learning rate to 0.01 and the hidden dimension to 256.
   \item \textit{Player2Vec.} Attributed heterogeneous information network proposed in~\cite{zhang2019key}. We set the same parameters as the FdGars model.
   \item \textit{Semi-GNN.} A semi-supervised graph attentive network for financial fraud detection proposed in~\cite{wang2019semi}. We set the learning rate to 0.001.
   \item \textit{GraphSAGE.} The inductive graph learning model proposed in~\cite{hamilton2017inductive}. We set the embedding dimension to 128.
   \item \textit{GraphConsis.} The GNN-based model tackling the inconsistency problem, proposed in~\cite{liu2020alleviating}. We used the default parameters suggested by the original paper.
   \item \textit{CARE-GNN.} The GNN-based model tackling fraud detection on a relational graph~\cite{dou2020enhancing}. We used default parameters from the original paper.
   \item \textit{PC-GNN.} A GNN-based model remedying the class imbalance problem, proposed in~\cite{liu2021pick}. We used the default parameters from the original paper.
   \item \textbf{GTAN.} The proposed gated temporal attention network model.\footnote{The sources of our proposed method GTAN will be available at \url{https://github.com/finint/antifraud}.} We also evaluate two variants of our model, GTAN-A and GTAN-R, in which the temporal graph attention component and risk embedding component are not considered, respectively. We set the batch size to 256, the learning rate to 0.0003, the input dropout ratio to 0.2, the number of heads to 4, the hidden dimension $d$ to 256, and train the model with the Adam optimizer for 100 epochs with early stopping.
\end{itemize}

\subsubsection{Evaluation Metrics}
We evaluate the experimental results on credit card fraud detection and opinion fraud datasets by the area under the ROC curve (AUC), macro average of F1 score (F1-macro), and averaged precision (AP), which are calculated as follows:

We count the number of True Positive $N_{TP}$ (i.e. correct identification of positive labels), False Positive $N_{FP}$ (i.e., incorrect identification of positive labels), and False Negatives $N_{FN}$ (i.e., incorrect identification of negative labels).
Then, F1 score and AP as formulated with ${F1}_{macro} =\frac{1}{l} \sum_{i=1}^{l} \frac{2 \times P_{i} \times R_{i}}{P_{i}+R_{i}}$ and $AP = \sum_{i=1}^{l}(R_i-R_{i-1})P_i$, where $P_i = N_{TP} / (N_{TP} + N_{FP})$ and $R_i = N_{TP} / (N_{TP} + N_{FN})$. We also report the AUC\footnote{\url{https://scikit-learn.org/stable/modules/generated/sklearn.metrics.auc.html}} in our experiments.%

\begin{figure}
\centerline{\includegraphics[width=1.0\linewidth]{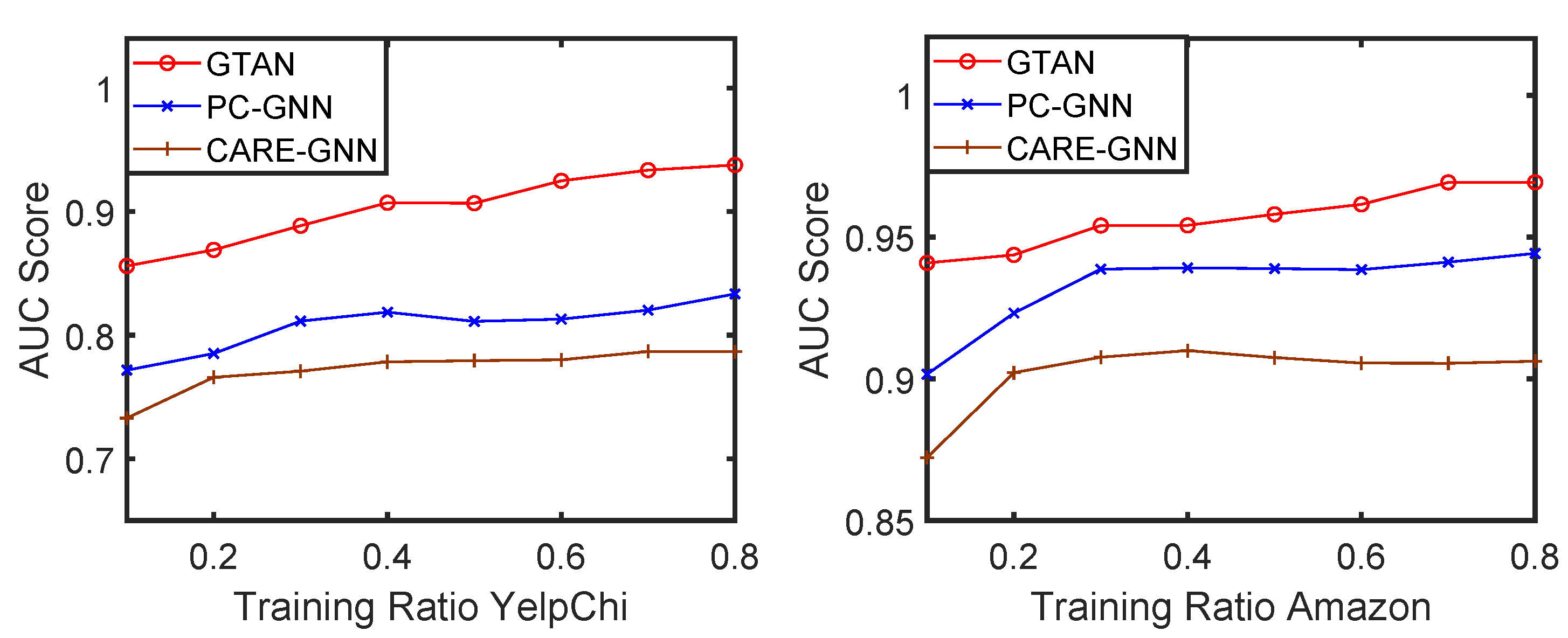}}
\vspace{-0pt}
\caption{The result of semi-supervised experiments with different ratios of labeled training data.}\label{fig:semi}
\vspace{-0pt}
\end{figure}

\begin{figure*}
\centerline{\includegraphics[width=1.0\linewidth]{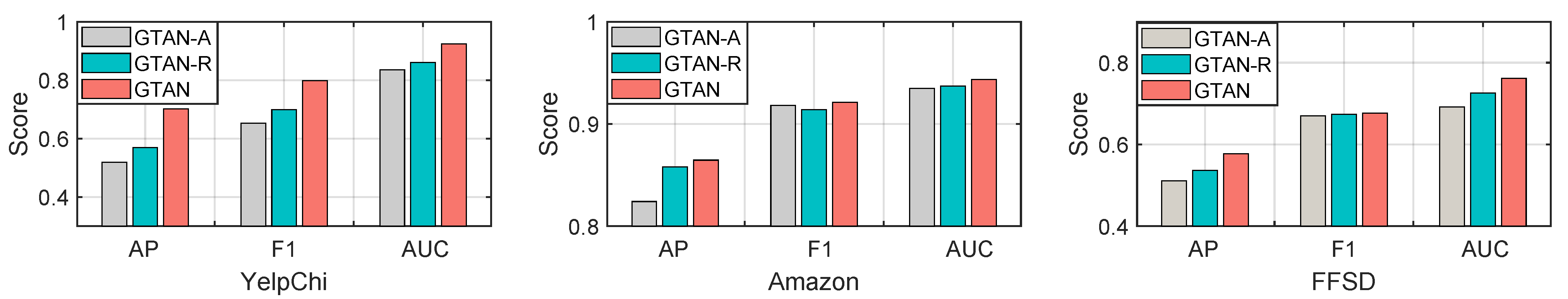}}
\caption{The ablation study results on three datasets. Gray bars represent the GTAN-A variant, blue bars represent the GTAN-R variant, and red bars represent the GTAN model.}\label{fig:aba}
\vspace{-0pt}
\end{figure*}

\subsection{Fraud Detection Performance}
In YelpChi and Amazon datasets, we set the ratio of training to testing as 2:3. In the FFSD dataset, transactions of the first $7$ months are used as training data, and then we detect fraud transactions in the following $3$ months (August, September, and October of 2021). We repeat the experiments ten times for each method and show the average performance in Table~\ref{tab:result}. $^{*}$ denotes that the improvements are statistically significant for $p<0.01$ according to the paired t-test.

The first five rows of Table \ref{tab:result} report the results of some classic graph-based methods, including GEM, Player2Vec, FdGars, Semi-GNN, and GraphSASE. It is clear that the result of GEM is not satisfactory, showing the limitation of a shallow model in addressing the complex fraud patterns. Player2Vec and FdGars improve the performance, partially due to the enlarged model capacity. Semi-GNN and GraphSAGE are close to each other and better than the first three methods. The result demonstrates the effectiveness of deep graph learning-based models in detecting fraud transactions.
PC-GNN achieves more competitive results by employing the transaction graphs in the learning process, which is considerably better than the above baselines.
The last row of Table \ref{tab:result} shows that our methods GTAN significantly outperforms all baselines with at least 10\%, 0.5\%, and 6\% AUC improvements in three datasets, respectively. Besides, our proposed GTAN also outperforms other baselines with at least 27\%, 2.9\%, and 12.8\% AP improvements in three datasets, respectively, which strongly proves the effectiveness of employing a semi-supervised temporal graph attention network for fraud detection.

\subsection{Semi-supervised Experiment}

To compare the capability of semi-supervised learning, we further set the ratio of the training set to the whole dataset to different values. For brevity of the diagrams, we select the two most competitive baselines (i.e., PC-GNN and CARE-GNN) for the following semi-supervised experiments. We vary the percentages of nodes used for training from 10\% to 80\% with an incremental of 10\% for eight sets of experiments, with the remaining nodes as the test set in each set of experiments. We perform experiments on the YelpChi and Amazon datasets since they are fully annotated, which allows us to vary the ratio of labeled data in a wide range. The experimental results are shown in Figure~\ref{fig:semi}.

On the YelpChi dataset, we can observe that GTAN always has the best performance under different training ratios. At the same time, in scenarios with little labeled data (10\% training ratio), GTAN still performs well. As the number of labeled data increases, there is a steady improvement in the performance of GTAN.
On the Amazon dataset, we can also observe that GTAN always has the best performance under different training ratios. Compared with the YelpChi dataset, the GTAN model on the Amazon dataset is less sensitive when changing the training ratio with no more than 2\% variations in the AUC. This fully demonstrates that GTAN can achieve good performance even when there is a small portion of labeled data available (i.e., as low as 10\%).
Therefore, we conclude that the GTAN model is robust to training ratio changes and consistently outperforms PC-GNN and CARE-GNN,
which shows the superiority of GTAN model in semi-supervised learning.

\begin{figure}[t]
\centerline{\includegraphics[width=1.0\linewidth]{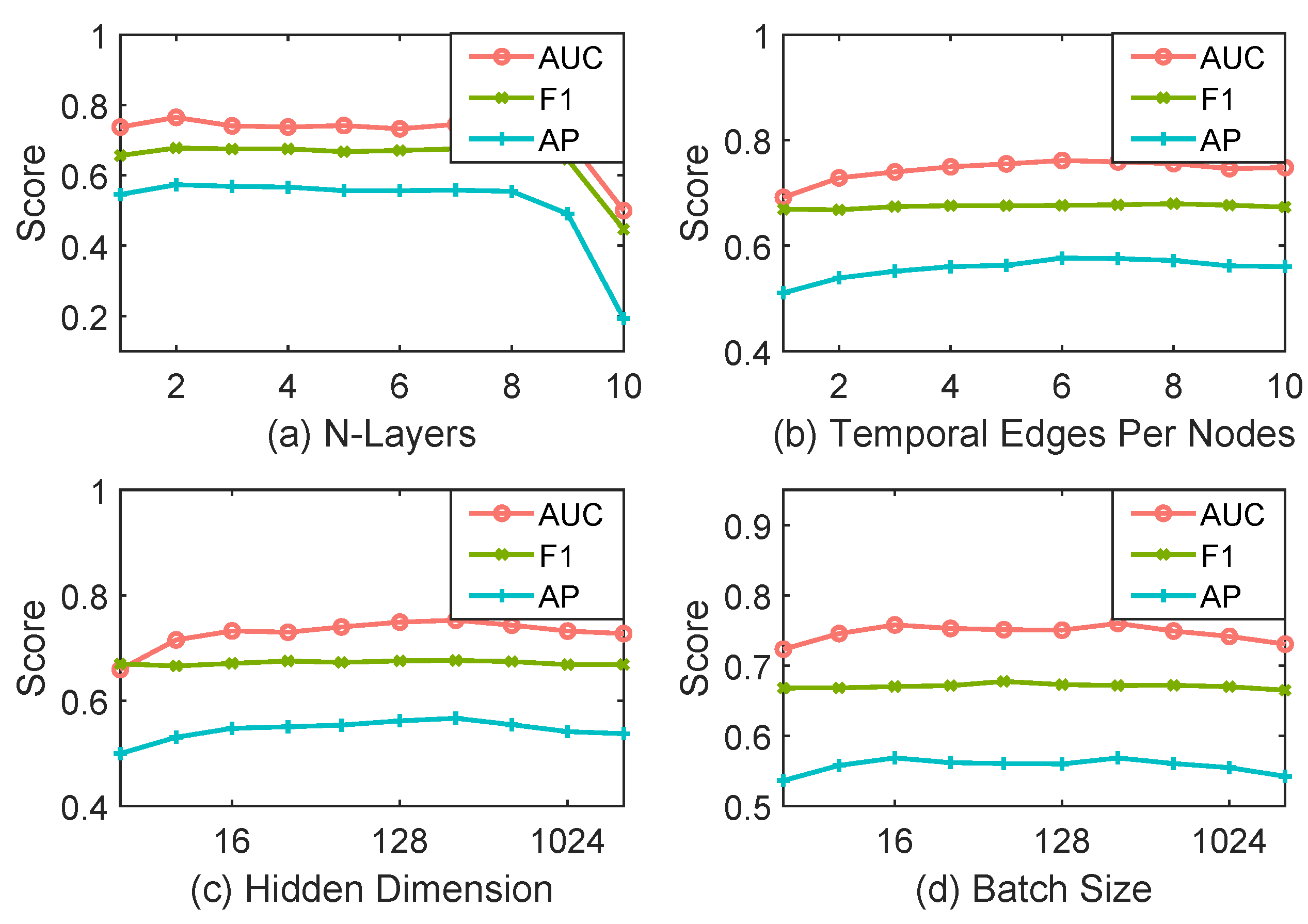}}
\vspace{-0pt}
\caption{Parameter sensitivity analysis with respect to (a) the number of GNN layers; (b) the number of temporal edges per node; (c) hidden dimension; and (d) the batch size.
}\label{fig:Para-Sens}
\vspace{-10pt}
\end{figure}

\subsection{Ablation Study}
The proposed model contains some key components, and we verify their effectiveness by ablating each component, respectively. Specifically, we evaluate two variants, namely GTAN-A and GTAN-R.
In the GTAN-A model, we remove the TGAT component, and the central node aggregates messages directly collected from neighboring nodes. In this case, we obtain the transaction embeddings from all the neighbor embeddings with the same weight instead of adaptively adjusting the weights of neighbor nodes. In the GTAN-R model, we remove the risk embedding component and only use the original node attributes $\mathbf{X}$.

The grey bars in Figure~\ref{fig:aba} show the removal of the attention mechanism has the greatest impact on the accuracy metrics of the model. This proves that the reweighting of temporal transaction neighbors from the temporal graph attention mechanism is effective in our graph-based method.
Besides, the green bars in Figure~\ref{fig:aba} gets the second-highest scores compared with GTAN, which proves that the risk embedding is effective in modeling credit card fraud patterns from transaction risk propagation.
In summary, removing either component deteriorates the performance of GTAN, which proves that the temporal graph attention mechanism and risk embedding are both effective in graph-based fraud detection.

\subsection{Parameter Sensitivity}
In this section, we study the model parameter sensitivity by varying the depth of temporal graph attention layers, the number of temporal edges per node, the hidden dimension, and the batch size. The experimental results in the fraud detection dataset are reported in Figure~\ref{fig:Para-Sens}.

We vary the depth of temporal graph attention layers from 1 to 10. As shown in Figure~\ref{fig:Para-Sens}(a), our model's performance remains stable with up to $8$ GNN layers. With deeper layers, our model tends to aggregate the temporal information from larger neighborhoods. Our model performs the best with two GNN layers when the AUC and AP reach the peak; therefore, we set the default depth to $2$. The performance is degraded if we keep on increasing the depth of TGAT layers. The reason might be that deeper GNNs result in over-smoothing~\cite{Zhao2020PairNormTO} in transaction embeddings.
Figure~\ref{fig:Para-Sens}(b) shows that when we increase the number of temporal edges per node from $1$ to $10$, our proposed model could consider neighbors in a wider range. Besides, our model requires at least $2$ neighbors to learn graph-based transaction embeddings, and it reaches peak performance when the number of edges is $6$.
Beyond that, an excessive increase in the number of edges in the graph is not beneficial to the model's accuracy. Figure~\ref{fig:Para-Sens}(c) shows that when we increase the hidden dimensions from $4$ to $2048$, our model maintains stable model performance and reaches a relative performance peak at $256$. Figure~\ref{fig:Para-Sens}(d) shows that our model performs best when the batch size is set as $64$ or $256$. Considering the training efficiency of the model, we set the batch size as $256$. Generally, for values ranging from $16$ to $512$, the model is not sensitive to the hidden dimension and batch size, with less than 3\% variations in the AUC.

\section{Conclusion}
\label{sec:con}

In this paper, we studied an important real-world problem of credit card fraud detection.
Considering that the labeling of fraud transactions is time-consuming and cost-expensive, we proposed an effective semi-supervised credit card fraud detection method by modeling data with temporal transaction graphs and developing attribute-driven gated temporal attention networks. Considering the ubiquitous categorical attributes and human-annotated labels, we proposed an attribute representation and risk propagation mechanism to model the fraud patterns accurately. The comprehensive experiments demonstrated the superiority of our proposed methods in three fraud detection datasets compared with other baselines. Semi-supervised experiments demonstrate the excellent fraud detection performance of our model with only a tiny proportion of manually annotated data. Our approach has been deployed in a transaction fraud analysis system. In the future, we will explore to study the temporal fraud patterns and risk-propagation fraud patterns in an effective and efficient way.

\section{Acknowledgments}
This work was supported by the National Science Fundation of China, 62102287, the Shanghai Science and Technology Innovation Action Plan Project (22511100700). Ling Chen was supported by ARC DP210101347.

\bibliography{aaai23}

\end{document}